\documentclass[conference]{IEEEtran}
\IEEEoverridecommandlockouts
\usepackage{cite}
\usepackage{amsmath,amssymb,amsfonts}
\usepackage{algorithmic}
\usepackage{graphicx}
\usepackage{textcomp}
\usepackage{xcolor}

\usepackage{svg}

\usepackage{subcaption}

\usepackage{scalerel}
\usepackage{tikz}
\usetikzlibrary{svg.path}

\definecolor{orcidlogocol}{HTML}{A6CE39}
\tikzset{
  orcidlogo/.pic={
    \fill[orcidlogocol] svg{M256,128c0,70.7-57.3,128-128,128C57.3,256,0,198.7,0,128C0,57.3,57.3,0,128,0C198.7,0,256,57.3,256,128z};
    \fill[white] svg{M86.3,186.2H70.9V79.1h15.4v48.4V186.2z}
                 svg{M108.9,79.1h41.6c39.6,0,57,28.3,57,53.6c0,27.5-21.5,53.6-56.8,53.6h-41.8V79.1z M124.3,172.4h24.5c34.9,0,42.9-26.5,42.9-39.7c0-21.5-13.7-39.7-43.7-39.7h-23.7V172.4z}
                 svg{M88.7,56.8c0,5.5-4.5,10.1-10.1,10.1c-5.6,0-10.1-4.6-10.1-10.1c0-5.6,4.5-10.1,10.1-10.1C84.2,46.7,88.7,51.3,88.7,56.8z};
  }
}

\newcommand\orcidicon[1]{\href{https://orcid.org/#1}{\mbox{\scalerel*{
\begin{tikzpicture}[yscale=-1,transform shape]
\pic{orcidlogo};
\end{tikzpicture}
}{|}}}}

\usepackage[hidelinks]{hyperref}
\def\BibTeX{{\rm B\kern-.05em{\sc i\kern-.025em b}\kern-.08em
    T\kern-.1667em\lower.7ex\hbox{E}\kern-.125emX}}
\makeatletter
\def\endthebibliography{%
  \def\@noitemerr{\@latex@warning{Empty `thebibliography' environment}}%
  \endlist
}
\makeatother
\begin{document}

% \title{Coherent Radiation Rendering of Polygonal Apertures\\
% % {\footnotesize \textsuperscript{*}Note: Sub-titles are not captured in Xplore and
% % should not be used}
% % \thanks{Identify applicable funding agency here. If none, delete this.}
% }
\title{Autocorrelation, Wigner and Ambiguity Transforms on Polygons for Coherent Radiation Rendering\\
}

\author{\IEEEauthorblockN{1\textsuperscript{st} Jacob Mackay}
\IEEEauthorblockA{\textit{Australian Centre for Field Robotics} \\
\textit{The University of Sydney}\\
Sydney, Australia \\
j.mackay@acfr.usyd.edu.au \orcidicon{0000-0003-3983-737X}}
\and
\IEEEauthorblockN{2\textsuperscript{nd} David Johnson}
\IEEEauthorblockA{\textit{Australian Centre for Field Robotics} \\
\textit{The University of Sydney}\\
Sydney, Australia \\
d.johnson@acfr.usyd.edu.au}
\and
\IEEEauthorblockN{3\textsuperscript{rd} Graham Brooker}
\IEEEauthorblockA{\textit{Australian Centre for Field Robotics} \\
\textit{The University of Sydney}\\
Sydney, Australia \\
g.brooker@acfr.usyd.edu.au}
}

\maketitle

\begin{abstract}
Simulating the radar illumination of large scenes generally relies on a geometric model of light transport which largely ignores prominent wave effects. This can be remedied through coherence ray-tracing, but this requires the Wigner transform of the aperture. This diffraction function has been historically difficult to generate, and is relevant in the fields of optics, holography, synchrotron-radiation, quantum systems and radar. In this paper we provide the Wigner transform of arbitrary polygons through geometric transforms and the Stokes Fourier transform; and display its use in Monte-Carlo rendering.
\end{abstract}


\section{Introduction}
\label{sec:intro}
The modelling and simulation of radar signal propagation can allow for performance predictions, detection algorithm development and scene reconstruction. Conventional ray-based light transport allows for fast simulation in large environments but doesn't account for the wave nature of light. For large-scale optical renders this is usually inconsequential, however at radar wavelengths or small optical scales wave effects become very prominent. One such effect of interest is the diffraction pattern of transmit apertures.

As outlined in \cite{Zernike1938, Walther1968, Wolf1978}, the radiation pattern of an aperture can be described using the coherence relationship. This yields the Wigner function, a bilinear spectral radiance distribution. Bastiaans \cite{Bastiaans1979a} outlined a framework in which this distribution can be transported along geometric rays, and many authors since have studied the propagation of these and similar distributions \cite{Alonso2011, Creagh2020}. In \cite{ZhengyunZhang2009}, an explicit connection between the Wigner function and the geometric light-field was analysed. This function was implemented for optical renderings as a diffraction shader by
% \cite{Oh2010, Cuypers2012}.
\cite{Cuypers2012}.
In \cite{Mackay2021}, the Wigner function was used for radar rendering including phased arrays, but only with simple elements. A result for a circular aperture was presented in \cite{Bastiaans1996} and further expanded by \cite{Mout2018}.

The Wigner transform has been widely used for phase-space modelling in the fields of optics, holography, synchrotron-radiation, quantum systems and radar. Use of this formulation has been mostly restricted to simple geometries, as the computation of this function is both analytically and numerically difficult due to the high dimensionality, inseparability and sampling densities required. Inspired by the work of \cite{Lee1983, Wuttke2017} on aperture transforms, we pursue an analytic formulation of the coherence distribution. In this paper we provide a polygonal form of the autocorrelation, Wigner and Ambiguity functions which is computationally tractable and physically accurate. We then use this to characterise the pattern of apertures, and integrate it into conventional renderers to verify our proposal. In Section \ref{sec:rad_model} we introduce the statistical radiometric models used to characterise the source of radiation. The main contributions of this paper are given in Section \ref{sec:method}, where we first introduce a motivating literature result, and then propose a solution to evaluating the radiation density functions. Section \ref{sec:interpret4D} links the Wigner function with observable radiometric quantities and provides visualisations of the 4D functions. The proposed method is compared with traditional diffraction in Section \ref{sec:results}, in addition to qualitative rendering. Concluding remarks are made in Section \ref{sec:conclusion}.

% \begin{figure}[htbp]
% \begin{figure}[t!]
% \centerline{\includegraphics[width=0.95\columnwidth]{./figures/vout_chev_O.png}}
% \caption{Volume render of two coherently radiating chevrons. A geometric Monte Carlo sampling method can be used with the Wigner transform of a transmitting element to easily model antenna diffraction. This can be integrated into ray-based renderers to model light transport appropriate to the radar context. The image is presented as a set of isosurfaces on a $\mathrm{dB}$ scale. Note the beam divergence and interference patterns fron the two transmitters.}
% \label{fig:volrender}
% \end{figure}

\begin{figure}[htbp]
% \centerline{\includegraphics[width=0.95\columnwidth]{./figures/aniout3.pdf}}
\centerline{\includesvg[width=0.95\columnwidth]{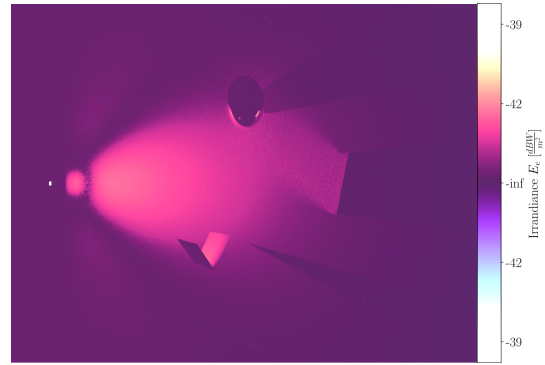}}
\caption{A coherent transmitter in a traditional rendering environment. The beam shape and divergence is clearly visible.}
\label{fig:render}
\end{figure}

\section{Statistical Radiometric Model}
\label{sec:rad_model}

In this section we present the radiometric model used to generate steady-state scenes illuminated by coherent sources. In this framework we retain the incoherent nature of ray traced rendering, but compute the coherence effect at the source rather than an observation point. In traditional rendering environments sources are considered either incoherent or point-like.
% This model takes in scene parameters such as geometries and reflectivities; transmitter geometry, illumination weight, efficiency, frequency, bandwidth and feed power; and receiver shape, sensitivity, and ADC characteristics. For the steady-state system, this model outputs the power received by a given receiver or test-radiometric camera.
As outlined in \cite{Walther1968}, the measurable signal is the power of a complex-valued scalar field originating from a transmitter. It is well known from the Wiener-Khinchin theorem that the energy spectral density (ESD) and autocorrelation (ACF) of the field form a Fourier dual pair with respect to the lag variable, providing the gain over a spectrum of spatial frequencies. If instead we consider the Fourier transform of the complex scalar field with respect to position, we are led to Fraunhofer diffraction \cite{Born2013}. This is valid in the far-field, however many real scenarios include near-field effects with a spatially varying spectral density. The field at the aperture of the transmitter can be modelled as a complex scalar field $u\left(\mathbf{x}\right)$, or statistically with autocorrelation functions
$R_{\mathbf{x_i x_j}} \left(\mathbf{x}_i, \mathbf{x}_j\right), R_{\mathbf{x}\boldsymbol\xi}\left(\mathbf{x}, \boldsymbol\xi\right)$, also known as the mutual power.

\begin{subequations}
    \begin{align}
        & u \left(\mathbf{x}\right) && = A\left(\mathbf{x}\right)e^{-2\pi \hat{\jmath}\boldsymbol\nu\cdot\mathbf{x}}
        \\
        & R_{\mathbf{x_i x_j}} \left(\mathbf{x}_i, \mathbf{x}_j\right) && = \left\langle u\left(\mathbf{x_i}\right)u^\dagger\left(\mathbf{x_j}\right)\right\rangle
        \\
        & R_{\mathbf{x}\boldsymbol\xi} \left(\mathbf{x}, \boldsymbol\xi\right) && = \left\langle u\left(\mathbf{x} + \boldsymbol\xi /2 \right)u^\dagger\left(\mathbf{x} - \boldsymbol\xi /2\right)\right\rangle \label{eqn:auto_lag}
    \end{align}
\end{subequations}

Where $\mathbf{x} = \left(\mathbf{x_i} + \mathbf{x_j}\right)/2$ is the position vector, $\boldsymbol\xi = \mathbf{x_i} - \mathbf{x_j}$ is the spatial lag, and $\boldsymbol\nu$ is the spatial frequency vector also known as the linear wavevector. The field $u$ is distributed over the aperture according to $A\left(\mathbf{x}\right)$. For simplicity, we drop the rapidly varying oscillation and simply take the root-mean-squared (rms) field for the expected value of the autocorrelation functions.
The ACF marginal distributions can be found by integrating over either the position or lag domains, and form Fourier pairs with what is typically called antenna gain.

\begin{subequations}
    \begin{align}
            & R_{\boldsymbol\xi}\left(\boldsymbol\xi\right) && = \int R_{\mathbf{x}\boldsymbol\xi} \left(\mathbf{x}, \boldsymbol\xi\right) \mathrm{d}\mathbf{x}
            \label{eqn:auto_pos_int}
            \\
            & R_{\mathbf{x}}\left(\mathbf{x}\right) && = \int R_{\mathbf{x}\boldsymbol\xi} \left(\mathbf{x}, \boldsymbol\xi\right) \mathrm{d}\boldsymbol\xi \label{eqn:auto_lag_int}
    \end{align}
\end{subequations}

The aperture gain as a function of the wavevector $\boldsymbol\nu$ or wavevector shift $\boldsymbol\upsilon$ can be found by taking the Fourier transform (FT) of the various distributions with respect to spatial lag $\boldsymbol\xi$ or position $\mathbf{x}$.

\begin{subequations}
    \begin{align}
            & F_{\boldsymbol\nu}\left(\boldsymbol\nu\right) && = \int u\left(\mathbf{x}\right) e^{-2\pi j \boldsymbol\nu \cdot \boldsymbol\xi} \mathrm{d}\boldsymbol\xi
            \label{eqn:ft}
            \\
            & W_{\mathbf{x}, \boldsymbol\nu}\left(\mathbf{x}, \boldsymbol\nu\right) && = \int R_{\mathbf{x}, \boldsymbol\xi} \left(\mathbf{x}, \boldsymbol\xi\right) e^{-2\pi j \boldsymbol\nu \cdot \boldsymbol\xi}\mathrm{d}\boldsymbol\xi
            \label{eqn:wf}
            \\
            & A_{\boldsymbol\upsilon, \boldsymbol\xi}\left(\boldsymbol\upsilon, \boldsymbol\xi\right) && = \int R_{\mathbf{x}, \boldsymbol\xi} \left(\mathbf{x}, \boldsymbol\xi\right) e^{-2\pi j \mathbf{x} \cdot \boldsymbol\xi}\mathrm{d}\mathbf{x}
            \label{eqn:af}
    \end{align}
\end{subequations}

Here $F_{\boldsymbol\nu}\left(\boldsymbol\nu\right)$ is the Fraunhofer field gain. $W_{\mathbf{x}, \boldsymbol\nu}\left(\mathbf{x}, \boldsymbol\nu\right)$, $A_{\boldsymbol\upsilon, \boldsymbol\xi}\left(\boldsymbol\upsilon, \boldsymbol\xi\right)$ are the position and lag power spectra, the famous Wigner function (WF) and Ambiguity function (AF) \cite{Bastiaans2009}. These provide the antenna gain as a function of position and wavevector, and lag and wavevector-shift.

% In this paper, we focus on the position wavevector power spectrum as it encodes the diffractive effects which are generally unnoticed in standard geometric optics.

% Do a section on kirchoff and fraunhofer diffraction
% In Australian language we don't usually do -, e.g. wavevector not wave-vector, cooperative not co-operative.

% \section{Manifold Transforms}
\section{Stokes Transforms}
\label{sec:method}

% The propagation of the positional wavevector ESD through a Wigner distribution has been the subject of many works \cite{}, however these have been confined to $1D$ functions, separable functions or simple distributions.

In this section, we address the problem of generating the positional density function. This is a particularly difficult problem due to the high dimensionality of the function, its inseparability across dimensions, and the generally high sampling resolution required for problems of interest. Whereas the Fast Fourier Transform (FFT) is a grid-based method, the Stokes transform is analytical and based on the geometry of the element. This allows for arbitrary sampling schemes and resolutions. We first outline a literature result which performs a Stokes Fourier transform (SFT). We then adapt this technique for the Wigner and Ambiguity transforms by finding appropriate slices of the autocorrelation function.

\subsection{Literature Result: Stokes Fourier Transform}
In \cite{Lee1983, Wuttke2017}, the authors describe the Fourier transform of coherently illuminated polygons and polyhedra. By considering a uniform illumination and Stokes' theorem, the transform can be found using a directed cyclic graph over the vertices.

% \begin{figure}[htbp]
\begin{figure}[t!]
    \centerline{\includesvg[width=0.85\columnwidth]{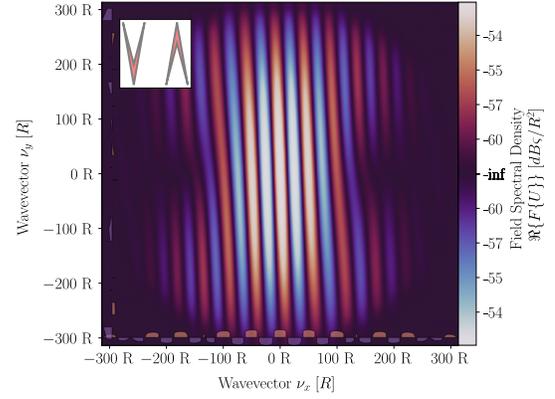}}
    \caption{Real part of Stokes Fourier transform of two coherent elements. This transform is valid for both convex and disjoint sets, remains analytical and is not bound by the grid requirements of the FFT. Here we display the function on a symmetric $\mathrm{dB}$ scale with wavevector axes down to $3.19~\mathrm{mm}$ ($94~\mathrm{GHz}$).}
    \label{fig:amp_spec}
\end{figure}

We begin by considering the Fourier transform of the polygon $\Gamma$ at frequency vector $\boldsymbol\nu \in SO\left(3, \mathbb{C}\right)$. The polygon $\Gamma$ is parameterised by the directed cyclic graph of vertices $\mathbf{V} \in E\left(3, \mathbb{R}\right)$ without intersection and with a winding number of $+1$ with respect to the normal $\hat{\mathbf{n}}$. The lag between vertices is $V'_i = V_i - V_j$ and the mean $\bar{V}_i = \frac{V_i + V_j}{2}$. The plane perpendicular, parallel and cross-parallel components of the frequency vector are given below.

% \begin{subequations}
%     \begin{align}
%         \boldsymbol\nu_{\perp} &= \left(\boldsymbol\nu \cdot \hat{\mathbf{n}}\right)\hat{\mathbf{n}}
%         \\
%         \boldsymbol\nu_{\parallel} &= \boldsymbol\nu - \boldsymbol\nu_{\perp}
%         \\
%         \boldsymbol\nu_{\times} &= \hat{\mathbf{n}} \times \boldsymbol\nu_{\parallel}
%     \end{align}
% \end{subequations}

\begin{equation}
    \boldsymbol\nu_{\perp} = \left(\boldsymbol\nu \cdot \hat{\mathbf{n}}\right)\hat{\mathbf{n}}
    ,\quad
    \boldsymbol\nu_{\parallel} = \boldsymbol\nu - \boldsymbol\nu_{\perp}
    ,\quad
    \boldsymbol\nu_{\times} = \hat{\mathbf{n}} \times \boldsymbol\nu_{\parallel}
\end{equation}

The Stokes Fourier transform (SFT) can be computed by iterating over the edges of the polygon.

\begin{equation}
        F\left(\boldsymbol \nu, \Gamma \right) = \frac{\boldsymbol \nu_{\times}^*}{2\pi \hat{\jmath} \left\| \boldsymbol \nu_{\parallel}\right\| _2^2}  \cdot \displaystyle \sum_{i=0}^{m} V_i'~\mathrm{sinc}\left(\boldsymbol \nu\cdot V_i'\right)e^{2\pi \hat{\jmath} \left(\boldsymbol \nu \cdot \bar{V_i}\right)}
        \label{eq:FVT}
\end{equation}

with the normalised sinc function $\mathrm{sinc}{\left(x\right)} = \frac{\sin{\left(\pi x\right)}}{\pi x}$. For a uniformly illuminated polyhedron, the SFT is found as the sum over sub-components. The transform for the three-dimensional figure $\Pi$ with $n$ faces indexed as $j$ is:

\begin{equation}
        F\left(\boldsymbol \nu, \Pi \right) = \frac{\boldsymbol \nu^*}{2\pi \hat{\jmath} \left\| \boldsymbol \nu\right\| _2^2}  \cdot \displaystyle \sum_{j=0}^{n} \hat{\mathbf{n}}_j F\left(\boldsymbol \nu, \Gamma_{j} \right)
\end{equation}

$\hat{\mathbf{n}}_j$ is the outward pointing normal of each sub-polygon, and we take $\boldsymbol\nu$ with respect to the frame of the polyhedron.

\subsubsection{1D Form}

Although trivial, we provide the SFT for a 1D figure. As the phase-space factors for 2D+ figures exist in a 4D+ space, the 1D version can be useful in visual debugging. Compared to the dense FFT, this form can also reduce the computational burden for sparse figures.

\begin{equation}
    F\left(\boldsymbol\nu, \mathrm{I} \right) = V'\mathrm{sinc}\left(\boldsymbol\nu\cdot V'\right)e^{2\pi \hat{\jmath} \left( \boldsymbol\nu \cdot \bar{V}\right)}
    \label{eq:FVT1D}
\end{equation}

\subsubsection{Multiple Radiators}

As a minor extension, we consider the effect of multiple coherently radiating elements. This is important for objects modelled as a mesh, for self-intersecting geometries such as the bow-tie antenna, or for multiple-input multiple-output (MIMO) and phased array antennas.

The transforms for the coherently radiating collections $\mathbf{I},~\boldsymbol\Gamma,~\boldsymbol\Pi$ are provided below. In the 1D case $\hat{\mathbf{p}}_i$ is a unit vector parallel to the edge.

\begin{subequations}
    \begin{align}
        F\left(\boldsymbol \nu, \mathbf{I}\right) &= \displaystyle \sum_{i=0}^{m} \left(1-\frac{\boldsymbol\nu^*}{\left\|\boldsymbol\nu\right\|_2} \cdot \hat{\mathbf{p}}_i\right) F\left(\boldsymbol\nu, \mathrm{I}_i \right)
        \\
        F\left(\boldsymbol \nu, \boldsymbol\Gamma \right) &= \frac{\boldsymbol \nu^*}{\left\| \boldsymbol \nu\right\| _2}  \cdot \displaystyle \sum_{j=0}^{n} \hat{\mathbf{n}}_j F\left(\boldsymbol \nu, \Gamma_{j} \right)
        \\
        F\left(\boldsymbol \nu, \boldsymbol\Pi \right) &= \frac{\boldsymbol \nu^*}{\left\| \boldsymbol \nu\right\| _2}  \cdot \displaystyle \sum_{k=0}^{o} \hat{\mathbf{n}}_k F\left(\boldsymbol \nu, \Pi_{k} \right)
    \end{align}
\end{subequations}

Figure \ref{fig:amp_spec} shows the result of the SFT on two chevrons down to a wavelength of $3.19~\mathrm{mm}$. Typically the magnitude squared of the FT is taken, however we simply display the real part of the transform to emphasise the complex nature of the function.

% If each element has a tunable phase-shift, we can drive the array in a direction $\mathbf{q} = \left[\sin{\theta_{az}}, \sin{\phi_{el}}\right]$ with respect to the array origin $\mathbf{s}_{tx} = \left(\mathbf{o}_{tx}, \mathbf{d}_{tx}\right)$. For an element with a geometric centre $\mathbf{o}_{e_k}$ and normal $\hat{\mathbf{n}}_{e_k}$, we have the transform from transmitter frame to element frame $\mathbf{H}$. The phase-shift is thus:
%
% \begin{equation}
%     \phi_{k}\left(\boldsymbol\nu\right) = \left\| \boldsymbol \nu\right\| _2\left(\mathbf{H}^{-1}\mathbf{o}_{e_k}\right)\cdot\mathbf{q}
% \end{equation}

\subsection{Autocorrelation Transforms}

We now consider the autocorrelation functions arising from coherent uniformly illuminated radiators. Although this function can easily be retrieved in a discrete framework, the sampling requirements can become prohibitive so we seek a closed form expression. The non-locality of the ACF further complicates sampling, but the region containing non-zero contributions can be found using the Minkowski automean.

% Recalling the definition \eqref{eqn:auto_lag}, for an $nD$ space the autocorrelation function is defined on a $2nD$ space. For spatial extents $\mathbf{D} = D_x, D_y, \dots , D_n$, and spatial frequency extents $\boldsymbol\Delta = \Delta_{\nu_x}, \Delta_{\nu_y}, \dots, \Delta_{\nu_n}$ we require $N = \displaystyle \prod \left(2\left\lceil\mathbf{D}_i \boldsymbol\Delta_i\right\rceil\right)^2$ samples for minimal Nyquist sampling.
% For example, a $10~mm\times10~mm$ plate radiating at $94~GHz=3.19~mm$ requires a minimum of $2^{12} = 4096$ dense samples.

% FIGURE OF SAMPLING GROWTH RATE. RESOLUTION, lambda/x VS SAMPLES

% We are thus motivated to explore other sampling regimes. Of interest is Monte-Carlo sampling as it is not bound by dimensionality \cite{}, however we require a closed-form expression for the value of the function at a given input.

\subsubsection{Full Autocorrelation \& Efficient Sampling}

\begin{figure}[t!]
% \centerline{\includesvg[width=0.95\columnwidth]{./figures/indmk.svg}}
\centerline{\includesvg[width=0.75\columnwidth]{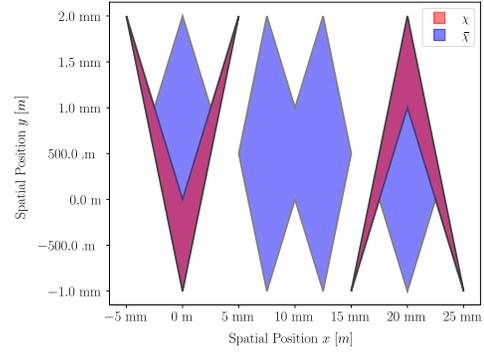}}
\caption{Two elements and their Minkowski automean. Red indicates the original radiators whilst the total region (red + blue) describes the support in position space of the autocorrelation function. Explicitly knowing this region can vastly reduce the sampling difficulty for non-compact or sparse figures as sampling volume is reduced as compared to the convex hull.}
\label{fig:indmk}
\end{figure}

The full autocorrelation function jointly describes the mutual power of a field in terms of position and lag, and evaluation of \eqref{eqn:auto_lag} is simple enough. When evaluating at an arbitrary position $\mathbf{x}$, it is worthwhile to know if the field will be non-zero. Noting a point in position-lag space $\left(\mathbf{x}, \boldsymbol\xi\right)$ corresponds to the two points in position space $\mathbf{x}_{i,j} = \mathbf{x} \pm \boldsymbol\xi/2$, the autocorrelation function is non-zero only if both $\mathbf{x}_{i,j}$ are within the mutual support of the original space. For convex radiators this support is the convex hull. However for non-convex or disjoint radiators it is more complex as illustrated in Figure \ref{fig:indmk}.

We define this mutual support as the set of position vectors where the autocorrelation function is non-zero, and call this the Minkowski automean. The Minkowski mean of two sets of position vectors $A, B$ is formed by taking the mean of each vector in $A$ with each vector in $B$. That is:

\begin{equation}
    \overline{A, B} = \left\{\frac{\mathbf{a}+\mathbf{b}}{2} ~|~ \mathbf{a} \in A, \mathbf{b} \in B\right\}
\end{equation}

The automean $\bar{A}$ is simply the result when $A$ and $B$ are identical. The mutual support for multiple radiators $\Gamma_{i,j,\dots, m}$ is found by first generating the set $\boldsymbol\Gamma = \textstyle\bigcup_{i=0}^{m} \Gamma_i$, then taking the automean $\bar{\boldsymbol\Gamma} = \overline{\boldsymbol\Gamma, \boldsymbol\Gamma}$.

% \begin{figure}[htbp]

\subsubsection{Autocorrelation Slices}
% If we start with a amplitude density, we get a density here. If we start with an amplitude, we get amplitude * cross section

The integrals \eqref{eqn:auto_pos_int}, \eqref{eqn:auto_lag_int} yield the total mutual power as a function of position or lag respectively, but can be difficult and expensive to compute. Similarly the Wigner and Ambiguity functions rely on integrals over a slice of the full ACF. A geometric interpretation allows slices to be found as products of the field with a transformed version of itself. Dropping the expectation value for brevity, a slice of the ACF at a position or lag can expressed:

% \begin{subequations}
%    \begin{align}
%            R_{\mathbf{x}, \boldsymbol\xi} \left(\mathbf{x}, : \right) &= \mathbf{T}^{ST}_{\left(\mathbf{x}\right)}U_{\left(:\right)}\mathbf{T}^{R}_{\left(\mathbf{x}\right)}U_{\left(:\right)}^H
%            \\
%            R_{\mathbf{x}, \boldsymbol\xi} \left(:, \boldsymbol\xi \right) &= \mathbf{T}^{T}_{\left(-\bar{\mathbf{x}}\right)} U_{\left(:\right)}\mathbf{T}^{T}_{\left(\boldsymbol\xi - \bar{\mathbf{x}}\right)}U_{\left(:\right)}^H
%    \end{align}
% \end{subequations}

% \begin{subequations}
%    \begin{align}
%            R_{\mathbf{x}, \boldsymbol\xi} \left(\mathbf{x}, \cdot \right) &= \mathbf{T}_{\left(-2\mathbf{x}\right)} \mathbf{S} u_{\left(\cdot \right)} \mathbf{T}_{\left(2\mathbf{x}\right)} \mathbf{R} u^*_{\left(\cdot\right)}
%            \\
%            R_{\mathbf{x}, \boldsymbol\xi} \left(\cdot, \boldsymbol\xi \right) &= \mathbf{T}_{\left(-\bar{\mathbf{x}}\right)} u_{\left(\cdot\right)} \mathbf{T}_{\left(\boldsymbol\xi - \bar{\mathbf{x}}\right)} u^*_{\left(\cdot\right)}
%    \end{align}
% \end{subequations}

\begin{subequations}
   \begin{align}
           R_{\mathbf{x}, \boldsymbol\xi} \left(\mathbf{x}, \cdot \right) &= \mathbf{M}_{\left(\mathbf{x}\right)}^{+} u_{\left(\cdot \right)} \mathbf{M}_{\left(\mathbf{x}\right)}^{-} u^*_{\left(\cdot\right)}
           \\
           R_{\mathbf{x}, \boldsymbol\xi} \left(\cdot, \boldsymbol\xi \right) &= \mathbf{T}_{\left(-\bar{\mathbf{x}}\right)} u_{\left(\cdot\right)} \mathbf{T}_{\left(\boldsymbol\xi - \bar{\mathbf{x}}\right)} u^*_{\left(\cdot\right)}
   \end{align}
\end{subequations}

with $\left(\cdot\right)$ denoting the entirety of a coordinate, $\mathbf{M}_{\left(\mathbf{x}\right)}^{+}$ the transformation matrix for scaling and transforming to the origin, $\mathbf{M}_{\left(\mathbf{x}\right)}^{-}$ the transformation for reflection about the origin, followed by scaling and transforming to the origin, $\mathbf{T}_{\left(\mathbf{a}\right)}$ for translation, and the mean geometric location of the field $\bar{\mathbf{x}}$.

\begin{equation}
    \mathbf{M}^{+}_{\left(\mathbf{x}\right)} = \left[\begin{array}{@{}c|c@{}}
                    2\mathbf{I} & -\mathbf{x} \\ \hline
                    0 & 1
                \end{array}\right]
    ,\quad
    \mathbf{M}^{-}_{\left(\mathbf{x}\right)} = \left[\begin{array}{@{}c|c@{}}
                    -2\mathbf{I} & \mathbf{x} \\ \hline
                    0 & 1
                \end{array}\right]
    ,\quad
    \mathbf{T}_{\left(\mathbf{a}\right)} = \left[\begin{array}{@{}c|c@{}}
                    \mathbf{I} & \mathbf{a} \\ \hline
                    0 & 1
                \end{array}\right]
\end{equation}

For a uniform illumination, the field $u$ can be considered an indicator function defined over the set $\Gamma$, with a mutual support $\bar{\Gamma}$. For a point in position space $\mathbf{x}$, we find the set $\Gamma'_{\boldsymbol\xi}\left(\mathbf{x}, \cdot \right)$ spanning lag space. Similarly a point in lag space $\boldsymbol\xi$ has the set $\Gamma'_{\mathbf{x}}\left(\cdot, \boldsymbol\xi\right)$.

% \begin{subequations}
%     \begin{align}
%         \chi'_{\boldsymbol\xi}\left(\mathbf{x}\right) &= \mathbf{T}^{ST}_{\left(\mathbf{x}\right)}\left(\chi \cap  \left(\mathbf{T}^{R}_{\left(\mathbf{x}\right)} \chi\right)\right)
%         \\
%         \chi'_{\mathbf{x}}\left(\boldsymbol\xi\right) &= \left(\mathbf{T}^{T}_{\left(-\bar{\mathbf{x}}\right)}\chi\right) \cap  \left(\mathbf{T}^{T}_{\left(\boldsymbol\xi-\bar{\mathbf{x}}\right)} \chi\right)
%     \end{align}
% \end{subequations}

% \begin{subequations}
%     \begin{align}
%         \Gamma'_{\boldsymbol\xi}\left(\mathbf{x}\right) &= \mathbf{T}_{\left(-2\mathbf{x}\right)}\mathbf{S}\left(\Gamma \cap  \left(\mathbf{T}_{\left(2\mathbf{x}\right)}\mathbf{R} \Gamma\right)\right)
%         \\
%         \Gamma'_{\mathbf{x}}\left(\boldsymbol\xi\right) &= \left(\mathbf{T}_{\left(-\bar{\mathbf{x}}\right)}\Gamma\right) \cap  \left(\mathbf{T}_{\left(\boldsymbol\xi-\bar{\mathbf{x}}\right)} \Gamma\right)
%     \end{align}
% \end{subequations}

\begin{subequations}
    \begin{align}
        \Gamma'_{\boldsymbol\xi}\left(\mathbf{x}, \cdot\right) &= \left(\mathbf{M}_{\left(\mathbf{x}\right)}^{+}\Gamma\right) \cap  \left(\mathbf{M}_{\left(\mathbf{x}\right)}^{-} \Gamma\right)
        \\
        \Gamma'_{\mathbf{x}}\left(\cdot, \boldsymbol\xi\right) &= \left(\mathbf{T}_{\left(-\bar{\mathbf{x}}\right)}\Gamma\right) \cap  \left(\mathbf{T}_{\left(\boldsymbol\xi-\bar{\mathbf{x}}\right)} \Gamma\right)
    \end{align}
\end{subequations}

\begin{figure}[htbp]
    \centering
    \begin{subfigure}[b]{0.7\columnwidth}
         \centering
         \includesvg[width=\columnwidth]{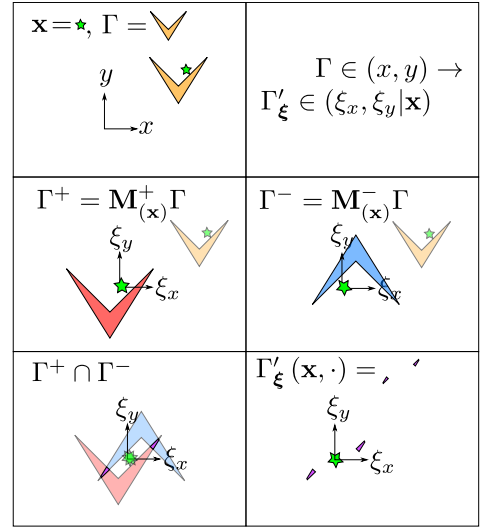}
         \caption{Transforms for lag-slice of autocorrelation function as a function of position.}
         \label{fig:auto_x}
     \end{subfigure}
     % \hfill
     \begin{subfigure}[b]{0.7\columnwidth}
          \centering
          \includesvg[width=\columnwidth]{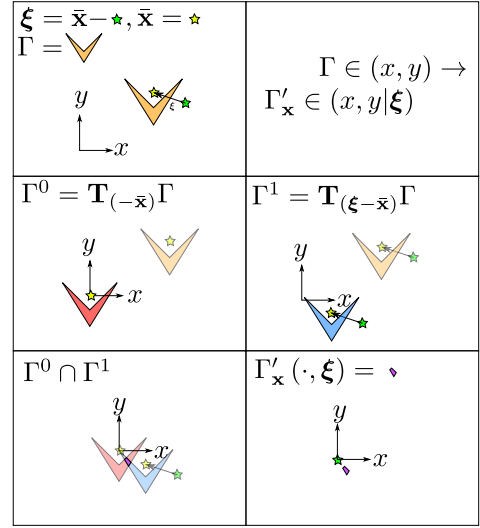}
          \caption{Transforms for position-slice of autocorrelation function as a function of lag.}
          \label{fig:auto_xi}
      \end{subfigure}
\caption{Geometric extraction of autocorrelation slices for a single element. Although the lag in (b) puts the sample outside the set, the autocorrelation function uses $\boldsymbol\xi/2$. From the slices $\Gamma'$, the Wigner and Ambiguity transforms can be evaluated. This method is not limited by scale or resolution, but a high model complexity could hamper intersection calculations.}
\label{fig:xforms}
\end{figure}

Figure \ref{fig:xforms} visually illustrates the various transforms for extracting slices in both position and lag space.

% Finding the marginals now corresponds to finding the measures of $\chi'_{\boldsymbol\xi}$, $\chi'_{\mathbf{x}}$, which for a polygonal figure is simply the area. The autocorrelation power cross-section at a given position or lag becomes:
%
% \begin{subequations}
%     \begin{align}
%         R_{\mathbf{x}} \left(\mathbf{x}\right) &= U^2\int_{\chi'_{\boldsymbol\xi}\left(\mathbf{x}\right)}\mathrm{d}\boldsymbol\xi
%         \\
%         R_{\boldsymbol\xi} \left(\boldsymbol\xi \right) &= U^2\int_{\chi'_{\mathbf{x}}\left(\boldsymbol\xi\right)}\mathrm{d}\mathbf{x}
%     \end{align}
% \end{subequations}

% \begin{figure}[htbp]
%     \centering
%     \begin{subfigure}[b]{0.9\columnwidth}
%          \centering
%          \includesvg[width=\columnwidth]{./figures/auto_x.svg}
%          \caption{}
%          \label{fig:auto_x}
%      \end{subfigure}
%      \hfill
%      \begin{subfigure}[b]{0.9\columnwidth}
%           \centering
%           \includesvg[width=\columnwidth]{./figures/auto_xi.svg}
%           \caption{}
%           \label{fig:auto_xi}
%       \end{subfigure}
% \caption{FIGURE OF ACF MARGINALS IN LAG AND POS. AGAIN, ROTATED TRIANGLES. ALSO SHOW THE CALCULATED.}
% \label{fig:indmk}
% \end{figure}

% I Guess the fractional transform comes from slices on other directions. I can write something else about this later.

\subsection{Stokes Wigner Transform}

With slices of the autocorrelation function at locations $\mathbf{x}$ spanning lags $\boldsymbol\xi$, we can now evaluate the Wigner transform \eqref{eqn:wf} using the Stokes Fourier transform \eqref{eq:FVT} for a specific wavevector $\boldsymbol\nu$.

\begin{equation}
        W_{\mathbf{x},\boldsymbol\nu}\left(\mathbf{x}, \boldsymbol\nu\right) = F\left(\boldsymbol\nu,\Gamma'_{\boldsymbol\xi}\left(\mathbf{x}, \cdot \right)\right)
\end{equation}

\subsection{Stokes Ambiguity Transform}

The Ambiguity function can be similarly found with autocorrelation slices spanning positions $\mathbf{x}$. For a specific wavevector shift $\boldsymbol\upsilon$ we have

\begin{equation}
        A_{\boldsymbol\upsilon, \boldsymbol\xi}\left(\boldsymbol\upsilon, \boldsymbol\xi\right) = F\left(\boldsymbol\upsilon,\Gamma'_{\mathbf{x}}\left(\cdot , \boldsymbol\xi\right)\right)
\end{equation}

\section{Interpreting 4D Functions}
\label{sec:interpret4D}

A common approach for presenting the 4D light field involves a tiled arrangement of function slices
% \cite{Dansereau2015, Wu2017}.
\cite{Wu2017}.
These can be difficult to interpret and noisy when a low number of samples are taken. An alternative approach is to display the marginal projections on orthogonal axes on a grid. This approach is visually simple, displays key features of the distribution, and has the advantage that the projections have meaningful physical interpretations.

\begin{figure}[htbp]
    \centering
    \begin{subfigure}[b]{0.9\columnwidth}
         \centering
         \includesvg[width=\columnwidth]{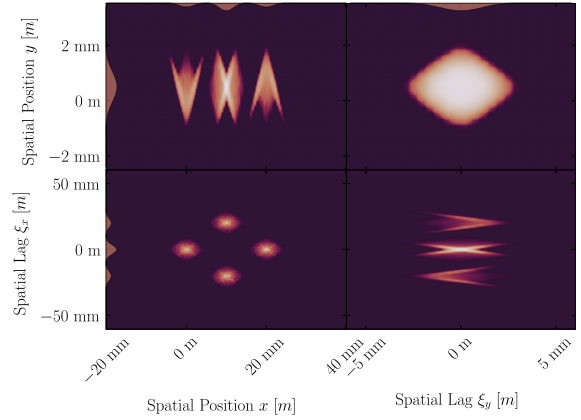}
         \caption{Projected autocorrelation function}
         \label{fig:auto_xxi}
     \end{subfigure}
     \hfill
     \begin{subfigure}[b]{0.9\columnwidth}
          \centering
          \includesvg[width=\columnwidth]{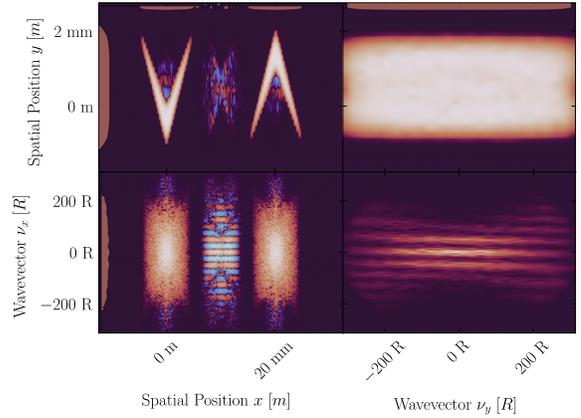}
          \caption{Projected Wigner function}
          \label{fig:w_xnu}
      \end{subfigure}
      \hfill
      \begin{subfigure}[b]{0.9\columnwidth}
           \centering
           \includesvg[width=\columnwidth]{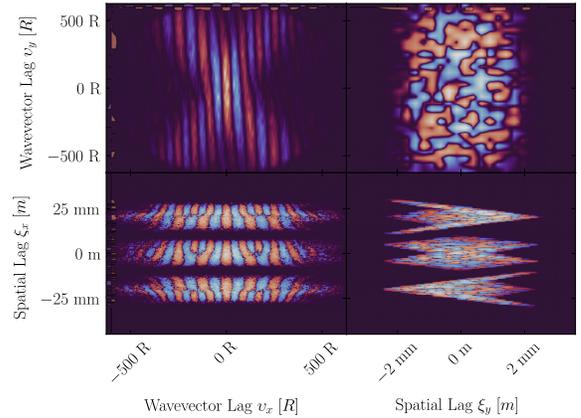}
           \caption{Projected Ambiguity function}
           \label{fig:a_upxi}
       \end{subfigure}
\caption{Projections of the real parts of the various bilinear functions. The diagonal tiles are projections onto complete domains such as position-position. For the Wigner function these correspond to measurables such as spectral intensity or radiant flux. These should be strictly real and positive semi-definite, however imperfect sampling prohibits this. The off diagonals are projections onto hybrid domains such as position-wavevector.}
\label{fig:projs}
\end{figure}

In Figure \ref{fig:projs} we display the projections of the autocorrelation, Wigner and Ambiguity functions. These are setup such that each image projection has a common orthogonal axis with its neighbour. The diagonal tiles contain measurable projections such as radiant intensity and radiant flux, whereas the off-diagonal elements contain interference components. It should be noted that these off-diagonals elements are full autocorrelation, Wigner and Ambiguity functions in their own right.

\subsection{Physical Interpretations}

With $u\left(\mathbf{x}\right)$ a complex scalar with amplitude $\left[\varsigma\right]$\footnote{For generality we give amplitude the units of $\left[\varsigma\right]$ which behaves as Voltage. We give power and power gain the units of $\left[\Sigma=\varsigma^2\right]$.}, we take the Wigner transform to find the diffractive effects. This yields a wavevector-spectral density with units of amplitude squared per square Raman\footnote{We denote units of inverse length as a Raman, so named after the physicist Chandrasekhara Venkata Raman. Just as: $\left[Hz\right] = \left[s^{-1}\right]$, $\left[R\right] = \left[m^{-1}\right]$.} $\left[\Sigma/R^2\right]$.
Radiance at the source for a particular wavelength $\lambda$ can be found through dividing by the spatial and wavevector measures. The radiance Wigner function $W_{L_{e, \Omega}}\left(\mathbf{x}, \hat{\mathbf{u}}\right)$ is now a function of the unit direction vector $\hat{\mathbf{u}}$ and has units $\left[\Sigma/\left(sr\cdot m^2\right)\right]$.

\begin{equation}
    W_{L_{e, \Omega}}\left(\mathbf{x}, \hat{\mathbf{u}}\right) = \frac{W\left(\mathbf{x}, \boldsymbol\nu\right)}{4\pi A \lambda^2}
\end{equation}

Through marginal projections, other familiar quantities can be found. The geometric flux vector $\mathbf{J}_{e}\left(\mathbf{x}\right)$ can be found by integrating the Wigner function with the normalised wavevector and area over the spatial frequency domain.
% Still maybe need to divide by 4pi. All of dnu should integrate to 4pilambda^2 assuming we're on so3
The spectral intensity of the surface, $I_{e, \Omega}$ can be found by integrating the Wigner function over the spatial domain. This does not include the dot-product of the emission, path and observation point normals. The total power $P_{e}$ can be found by integrating over the entire spatial-wavevector domain. %The total radiant flux $\Phi_{e}$ is found by integrating the radiative flux density over the spatial domain.

\begin{subequations}
    \begin{align}
        &\mathbf{J}_{e}\left(\mathbf{x}\right) &&= \int\frac{\boldsymbol\nu}{A\left\|\boldsymbol\nu\right\|_2} W\left(\mathbf{x}, \boldsymbol\nu\right)\mathrm{d}\boldsymbol\nu &\quad \left[\Sigma/m^2\right]
        \\
        &I_{e, \Omega}\left(\boldsymbol\nu\right) &&= \int W\left(\mathbf{x}, \boldsymbol\nu\right)\mathrm{d}\mathbf{x} &\quad \left[\Sigma/R^2\right]
        \\
        &P_{e} &&= \iint W_{L_{e, \Omega}}\left(\mathbf{x}, \boldsymbol\nu\right)\mathrm{d}\mathbf{x}\mathrm{d}\boldsymbol\nu &\quad \left[\Sigma\right]
    \end{align}
\end{subequations}

It should be noted that the Wigner function becomes negative in parts, violating the requirements for it to be a true radiance function. However, \cite{Friberg1979} shows that it is impossible for any radiance function to satisfy all physical requirements. It is shown in \cite{ZhengyunZhang2009}, that any measurement of this field described by Wigner diffraction which encompasses a volume satisfying the uncertainty principle is positive semi-definite. For more details and other physical interpretations such as effective width and beamwidth see \cite{Bastiaans2009}.
% Keep expanding for thesis sake

% Also investigate the manifold/spherical question

% Note that polarisation is explored in alonso
% Have a look at shperical illumination and non-uniform illumination
% Note about gaussian illumination and its extensive work

% Idea for inverse fourier transform...challenging as you need an analytic/polygonal slice of some domain.
% Perhaps polyganise, then use my proposed non-uniform approach?

% Quick version:
% Volume render

% Figures todo: Qualitative render
% Qualitative volume render wvt
% Qualitative volume render fdtd
% Qualitative volume render fft
% Qualitative volume render uniform + cosine
% Quantitative error vs (r or z)

\section{Results and Discussion}
\label{sec:results}

We demonstrate the coherent Wigner transmitter as a plugin to the Mitsuba2 renderer \cite{Nimier-David2019}. The radiometer is modelled as a pinhole camera sensitive to a narrowband, at $94~\mathrm{GHz}$ and each pixel corresponds to a small $\Delta\Omega$.

Following the Kajiya formulation of light transport
% \cite{Kajiya1986}
the power transported through a series of light paths is evaluated. We consider the radiance at a point in the world frame $\mathbf{x}'$ equipped with a normal $\hat{\mathbf{n}}'$ arising from direct illumination by the transmitter. A point $\mathbf{x}$ on the transmitter's mutual support $\bar{\Gamma}$ is sampled, and the direct path between the two is found $\mathbf{r} = \mathbf{x}' - \mathbf{x}$. Normalising $\mathbf{r}$ we retrieve the outgoing wavevector for a specified wavelength
 % $\boldsymbol\nu^{tx} = \frac{\hat{\mathbf{u}}^{tx}}{\lambda}$
 $\boldsymbol\nu = \frac{\mathbf{r}}{\lambda \left\|\mathbf{r}\right\|_2}$.
A simplified transport equation can be written for the Poynting flux at a world point arising from a single phase-space component of the coherent transmitter:

\begin{equation}
    \mathrm{d}\mathbf{S}\left(\mathbf{x}'\right) = W\left(\mathbf{x}, \boldsymbol\nu\right)  \frac{\left(\hat{\mathbf{n}} \cdot \hat{\mathbf{r}}\right)\hat{\mathbf{r}}}{\left\|\mathbf{r} \right\|_2^2} \frac{A}{\lambda^2}
    % L\left(\mathbf{x}_{w}\right) = W\left(\mathbf{x}_{tx}, \boldsymbol\nu^{tx}\right)  \frac{1}{4\pi \lambda^2 \left\|\mathbf{u} \right\|_2^2}\left(\hat{\mathbf{n}}_{tx} \cdot \hat{\mathbf{u}}\right)\left(\hat{\mathbf{n}}_{w}\cdot \hat{\mathbf{u}}\right)
    % L\left(\mathbf{x}_{w}\right) = W^L\left(\mathbf{x}_{tx}, \boldsymbol\nu^{tx}\right)  \frac{A}{\left\|\mathbf{u} \right\|_2^2}\left(\hat{\mathbf{n}}_{tx} \cdot \hat{\mathbf{u}}\right)\left(\hat{\mathbf{n}}_{w}\cdot \hat{\mathbf{u}}\right)
    \label{eq:radiance}
\end{equation}

% Where $A$ is the surface area of the transmitter.
A more detailed treatment of transport and diffraction of Wigner functions can be found in \cite{Bastiaans1979a, Creagh2020}. Figure \ref{fig:render} displays a simple scene rendered with an above-view radiometric camera and a small patch transmitter. The characteristic beam lobes can easily be seen in both elevation and azimuth. Some noise is present about the nulls, but this can be remedied with either sufficient sampling or a more robust strategy.

% \begin{figure}[htbp]
% \centerline{\includegraphics[width=0.8\columnwidth]{./figures/aniout3.pdf}}
% % \centerline{\includesvg[width=0.5\columnwidth]{./figures/aniout2.svg}}
% \caption{A coherent transmitter in a traditional rendering environment. The beam shape and divergence is clearly visible.}
% \label{fig:render}
% \end{figure}

% \begin{figure}[t!]
% \centerline{\includegraphics[width=0.8\columnwidth]{./figures/vout_chev_O.png}}
% \caption{Volume render of two coherently radiating chevrons. A geometric Monte Carlo sampling method can be used with the Wigner transform of a transmitting element to easily model antenna diffraction. This can be integrated into ray-based renderers to model light transport appropriate to the radar context. The image is presented as a set of isosurfaces on a $\mathrm{dB}$ scale. Note the beam divergence and interference patterns fron the two transmitters.}
% \label{fig:volrender}
% \end{figure}
\begin{figure}[t!]
\centerline{\includegraphics[width=0.77\columnwidth]{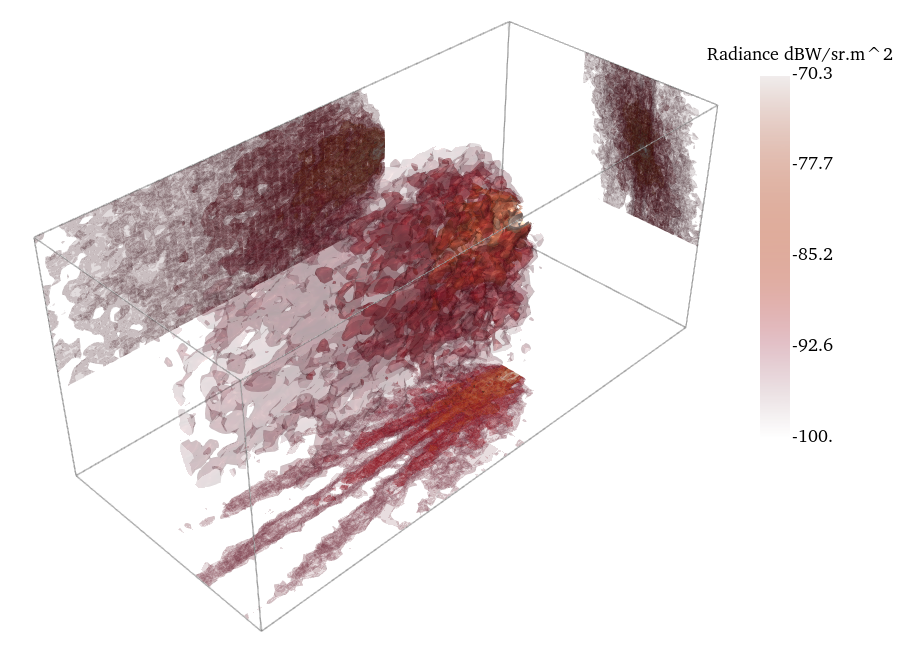}}
\caption{Volume render of two coherently radiating chevrons presented as a set of isosurfaces on a $\mathrm{dB}$ scale. Note the beam divergence and interference patterns from the two transmitters.}
\label{fig:volrender}
\end{figure}

\begin{figure}[htbp]
    \centering
    \begin{subfigure}[b]{0.95\columnwidth}
         \centering
         \includesvg[width=\columnwidth]{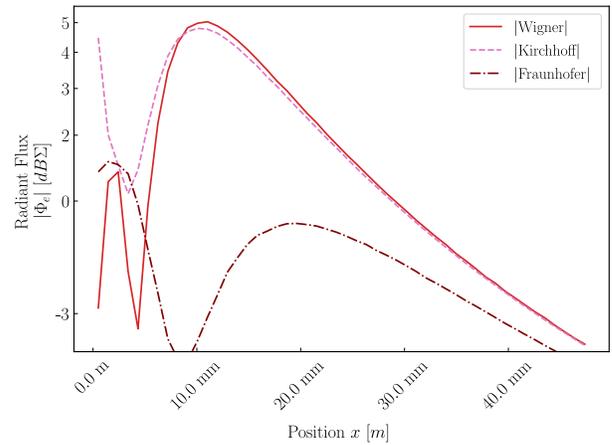}
         \caption{Flux as a function of distance along the boresight}
         \label{fig:line_powers}
     \end{subfigure}
     \hfill
     \begin{subfigure}[b]{\columnwidth}
          \centering
          \includesvg[width=\columnwidth]{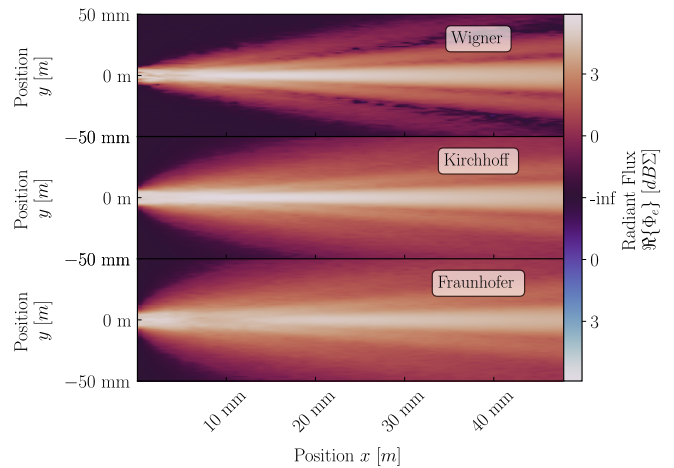}
          \caption{Flux in a plane containing the transmitter}
          \label{fig:cplane_powers}
      \end{subfigure}
\caption{Comparisons between Wigner diffraction and traditional diffraction equations. A simple square $10~\mathrm{mm}\times10~\mathrm{mm}$ transmitter at $94~\mathrm{GHz}$ was uniformly illuminated at unity. On boresight (a), the Wigner diffraction converges to Kirchhoff after $2\lambda$. Off-boresight Wigner results (b) show a strong response in the lobes, but weakened for higher angles due to the simplistic propagation method.}
\label{fig:compar}
\end{figure}

A volumetric render of the radiance field arising from the coherent chevrons is displayed in Figure \ref{fig:volrender} as a series of isosurfaces and orthogonal projections. An advantage of the sampling based approach is that the field can be evaluated at arbitrary resolution or within a subvolume of interest. Interpolating or kernel fitting can be used to upsample the continuous field. A quantitative rendering was performed in Figure \ref{fig:compar} to compare Wigner diffraction using \eqref{eq:radiance} with Kirchhoff and Fraunhofer diffraction. Having the analytical solution to the Wigner diffraction integral allows for straightforward sampling, whereas other diffraction integrals must be numerically evaluated and propagated.

% Compare to fdtd for both accuracy and time. Compare to fft and uniform.

% \begin{figure}[b!]
% % \begin{figure}[htbp]
% \centerline{\includegraphics[width=0.8\columnwidth]{./figures/vout_chev_O.png}}
% % \centerline{\includesvg[width=0.5\columnwidth]{./figures/aniout2.svg}}
% \caption{Volume rendering of the two chevrons. The radiance field displays the interference between the transmitters as a result of the accumulation of many plane wave samples.}
% \label{fig:volrender2}
% \end{figure}

% \subsection{Extensions to the Wigner Vertex Transform}
% In this section...
% \subsubsection{Non-Uniform Illumination}
%
% \subsubsection{Phased Arrays}

\section{Conclusion}
\label{sec:conclusion}

In summary, we propose a method to generate the autocorrelation, Wigner and Ambiguity functions for polygonal apertures. Doing so enables Monte-Carlo sampling and subsequent ray traced rendering of coherent transmitters whilst retaining wave effects in an incoherent sampling environment. This provides an additional approach to practical, fast and accurate rendering of radar scenes, including illumination from non-convex and disjoint transmitters. Future extensions accounting for variations in the complex amplitude of the incident field could allow this technique to address phased arrays and apertures with tapered illuminations.

% \section*{Acknowledgement}

\bibliography{./references/PGC.bib}

% \section{Appendix}
%
% Additional patterns
%
% \begin{itemize}
%     \item 1D single slit
%     \item 1D heterogeneous, offcentre double slit
%     \item 2D single rectangle
%     \item 2D Circle
%     \item 2D triangle
%     \item 2D heterogeneous, offcentre double slit
%     \item 2D V
%     \item 2D U
%     \item 2D intereoferometer graham
% \end{itemize}

\end{document}